\documentclass[journal,twoside,web]{ieeecolor}

\usepackage{etoolbox}
\makeatletter
\@ifundefined{color@begingroup}%
{\let\color@begingroup\relax
\let\color@endgroup\relax}{}%
\def\fix@ieeecolor@hbox#1{%
\hbox{\color@begingroup#1\color@endgroup}}
\patchcmd\@makecaption{\hbox}{\fix@ieeecolor@hbox}{}{\FAILED}
\patchcmd\@makecaption{\hbox}{\fix@ieeecolor@hbox}{}{\FAILED}

\usepackage{tmi}
\usepackage{cite}
\usepackage{amsmath,amssymb,amsfonts}
\usepackage{algorithmic}
\usepackage{graphicx}
\usepackage{textcomp}
\usepackage{algorithm}
\usepackage{graphics}
\usepackage{threeparttable}
\usepackage{color}
\usepackage[normalem]{ulem}
\usepackage{multirow}
\usepackage{float}
\usepackage{amsfonts}
\usepackage{bm}
\usepackage{array}
\usepackage{colortbl}
\usepackage{pifont}
\usepackage{diagbox}
\usepackage{rotating}
\usepackage{booktabs}
\usepackage{overpic}
\usepackage{makecell}
\usepackage{contour}
\usepackage{url}
\usepackage[colorlinks,linkcolor=blue,anchorcolor=blue,citecolor=blue,]{hyperref}

\newcommand{\tabref}[1]{Table \ref{#1}}

\newcommand{\figref}[1]{Fig. \ref{#1}}

\newcommand{\secref}[1]{Sec. \ref{#1}}

\newcommand{\blue}[1]{{\textcolor{black}{#1}}}

\def\ie{\emph{i.e.}}
\def\eg{\emph{e.g.}}

\def\etal{{\em et al.}}

\def\ourmodel{\emph{S$^2$Former-OR}}

\graphicspath{{./Imgs/}}
\DeclareGraphicsExtensions{.jpg,.pdf,.png}

\def\BibTeX{{\rm B\kern-.05em{\sc i\kern-.025em b}\kern-.08em
    T\kern-.1667em\lower.7ex\hbox{E}\kern-.125emX}}
\begin{document}
\title{S$^2$Former-OR: Single-Stage Bi-Modal Transformer for Scene Graph Generation in OR}
%\title{Single-stage Multi-view bi-modal Transformer for Scene Graph Generation in Operating Rooms}

\author{Jialun Pei, \IEEEmembership{Member, IEEE}, Diandian Guo, Jingyang Zhang, Manxi Lin, Yueming Jin, \IEEEmembership{Member, IEEE}, \\ Pheng-Ann Heng, \IEEEmembership{Senior Member, IEEE}
% \thanks{This work was supported in part by the XX Department of Commerce under Grant XXXXX. (Corresponding author: Yueming Jin)}
\thanks{J. Pei, D. Guo, J. Zhang, and P.-A. Heng are with the Department of Computer Science and Engineering, P.-A. Heng is also with the Institute of Medical Intelligence and XR, The Chinese University of Hong Kong, HKSAR, China. (e-mail: jialunpei@cuhk.edu.hk, guodiandian1998@gmail.com, {jyzhang, pheng}@cse.cuhk.edu.hk)}
\thanks{M. Lin is with the Department of Applied Mathematics and Computer Science, Technical University of Denmark, Kongens Lyngby, Denmark. (email: manli@dtu.dk)}
\thanks{Y. Jin is with the Department of Electrical and Computer
Engineering, National University of Singapore, Singapore.(e-mail:
ymjin@nus.edu.sg)}
\thanks{Jialun Pei and Diandian Guo contributed equally.}
\thanks{Corresponding author: Yueming Jin.}
}

\maketitle

\begin{abstract}
Scene graph generation (SGG) of surgical procedures is crucial in enhancing holistically cognitive intelligence in the operating room (OR). 
%%with far-reaching consequences for surgical workflow optimization and team cooperation. 
% However, previous works have primarily relied on the multi-stage learning that generates semantic scene graphs dependent on intermediate processes with pose estimation and object detection, which may compromise model efficiency and efficacy, also impose extra annotation burden.
\blue{However, previous works have primarily relied on multi-stage learning, where the generated semantic scene graphs depend on intermediate processes with pose estimation and object detection. 
This pipeline may potentially compromise the flexibility of learning multimodal representations, consequently constraining the overall effectiveness.}
In this study, we introduce a novel single-stage bi-modal transformer framework for SGG in the OR, termed \ourmodel, aimed to complementally leverage multi-view 2D scenes and 3D point clouds for SGG in an end-to-end manner. 
Concretely, our model embraces a View-Sync Transfusion scheme to encourage multi-view visual information interaction. Concurrently, a Geometry-Visual Cohesion operation is designed to integrate the synergic 2D semantic features into 3D point cloud features. % to obtain a unified representation. 
Moreover, based on the augmented feature, we propose a novel relation-sensitive transformer decoder that embeds dynamic entity-pair queries and relational trait priors, which enables the direct prediction of entity-pair relations for graph generation without intermediate steps. 
Extensive experiments have validated the superior SGG performance and lower computational cost of \ourmodel~on 4D-OR benchmark, compared with current OR-SGG methods, \eg, \blue{3 percentage points Precision increase} and 24.2M reduction in model parameters.
%0.3s reduction in per-scene inference speed.
We further compared our method with generic single-stage SGG methods with broader metrics for a comprehensive evaluation, with consistently better performance achieved.
\blue{Our source code can be made available at}: \url{https://github.com/PJLallen/S2Former-OR}.

\end{abstract}


\section{Introduction}\label{sec:introduction}
\IEEEPARstart{A}{s} medicine and information technology advance, the operating room (OR) has evolved into a highly complicated and technologically rich environment. This setting is characterized by high variability and dynamic, unpredictable scenarios~\cite{lalys2014surgical}. Surgical procedures in the modern OR are increasingly complex, necessitating intricate coordination among medical personnel, patients, and a multitude of devices~\cite{maier2017surgical}. An automatic and holistic comprehension of surgical procedures is critically essential for monitoring and optimizing surgical processes, scheduling surgeons, and enhancing coordination among surgical teams~\cite{kennedy2020computer}. To this end, holistically understanding OR can augment OR efficiency, improve patient safety, and enhance the quality of patient treatment~\cite{chang2021comprehensive}. Scene graph generation (SGG), a powerful semantic representation technique derived from the computer vision field, has shown great potential in capturing complex interactive relationships between entities in the 3D dynamic OR environment~\cite{chang2021comprehensive,ozsoy2023labrad}. Effective SGG facilitates a comprehensive understanding of the entire OR scene, thus highly desirable for promoting cognitive assistance for the next generation of operating intelligence.

%\IEEEPARstart{I}{n} the surgical process, a comprehensive understanding of specific surgical procedures is critically essential for supervising and optimizing technical processes through computer-aided interaction, which offers new avenues for enhancing surgical effectiveness. Scene understanding and context-awareness play a crucial role in the field of medical semantic scene analysis, especially in operating room (OR) scenarios~\cite{ozsoy20224d}. Scene graph generation (SGG), a powerful semantic representation of visual understanding, has shown great potential in capturing complex interactive relationships between entities and an overall overview of an OR setting~\cite{chang2021comprehensive,ozsoy2023labrad}. By recognizing rich semantic relationships between medical staff, patient, and instrument in the 3D dynamic surgical environment, SGG provides valuable guidance on holistic OR modeling and understanding, which yields versatile advantages for surgical procedures, such as surgeon decision-making~\cite{maier2022surgical}, patient outcomes~\cite{kennedy2020computer}, intra-operative supervision~\cite{bian2023motion}, and surgical workflow recognition~\cite{jin2021temporal}. 

\begin{figure}[t!]
\centering
\includegraphics[width=\linewidth]{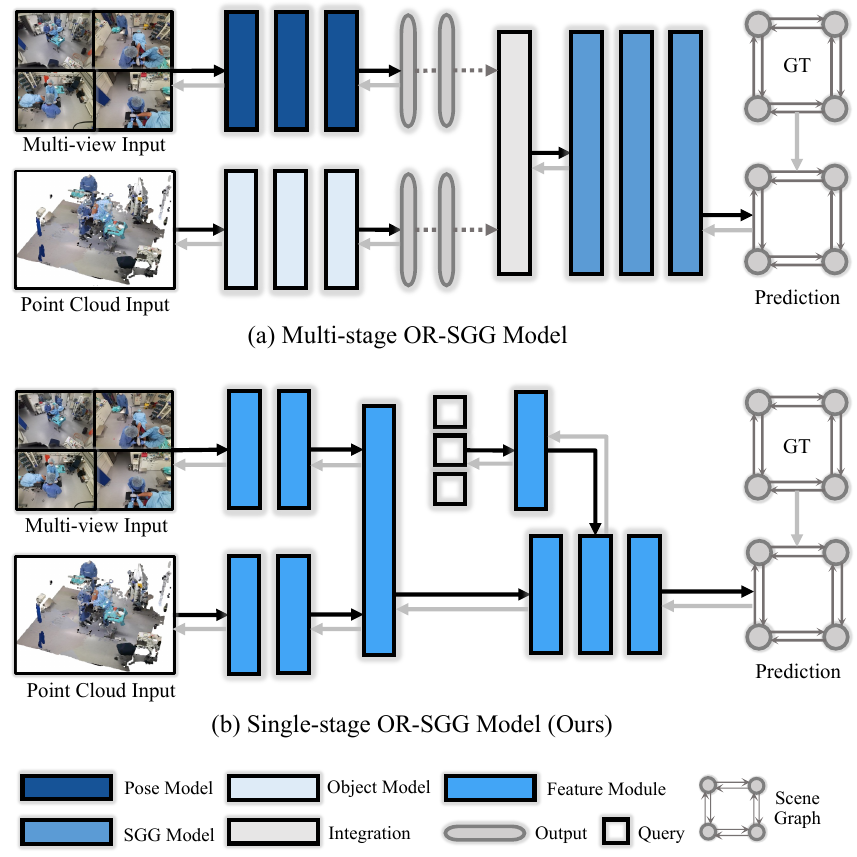}
\caption{\blue{Diagrams of multimodal architectures for scene graph generation in operating rooms. (a) Existing multi-stage model~\cite{ozsoy20224d}; (b) Our proposed single-stage model in an end-to-end manner. Black arrows represent forward inference while grey arrows represent backpropagation during the training phase.}}
\label{struc_com}
\end{figure}

To advance SGG methods for OR modeling, Özsoy \etal~\cite{ozsoy20224d} introduced the first OR-SGG dataset, termed 4D-OR, which contains multi-view video frames and corresponding 3D point clouds in the knee surgery theatre. 
Moreover, they proposed a multi-stage semantic SGG pipeline that provides 3D human pose estimation and object detection, followed by blending different forms of prediction labels and constructing a scene graph through a 3D SGG framework~\cite{wald2020learning}. 
Subsequently, LABRAD-OR~\cite{ozsoy2023labrad} proposed to enhance the multi-stage framework by incorporating temporal information into scene graphs. The constructed memory scene graphs can capture fine-grained relationships between entities in dynamic OR, achieving state-of-the-art results.
% However, existing methods rely on a multi-stage model training (shown in \figref{struc_com} (a)), typically facing the following challenges which may hinder its applicability in real-world deployment:
% i) In terms of efficiency, multi-stage approaches require step-wise processing, which imposes additional annotation burdens and computational complexity during model training. Meanwhile, the inherent structure of a multi-stage framework elongates inference time, potentially compromising the ability to detect abnormalities in OR promptly.
% %which brings additional training burden, increases computational complexity and potentially bottlenecks the performance; 
% % resulting in increased computational complexity and potentially leading to performance bottlenecks; 
% ii) In terms of precision, the intermediate representations generated at each stage may not fully retain the contextual information desired for scene modeling, consequently failing to capture the long-range dependencies within the scene and risking losing representational information.
\blue{Generally, existing OR-SGG methods based on multi-stage step-wise training (shown in~\figref{struc_com}(a)), typically confront the following challenges:
i) For multimodal scene modeling, the intermediate representations produced by each stage may be inadequate in preserving the contextual information desired for scene generating, consequently being suboptimal to capturing the long-range dependencies within the scene and constraining the adaptability of representation learning.
ii) The multi-stage pipeline requires step-by-step training of multiple subtasks, which may limit operational efficiency and hinder its deployment and application in realistic surgery.}
% More crucially, the multi-stage architecture may hinder its applicability to real-world tasks that require real-time performance.
%iii) For instance, generating scene graphs promptly in the practical surgical process can effectively control the appearance of abnormalities. Its application values are significantly undermined in a multi-stage setting, due to the terrific time cost. 
% If the input samples require multi-stage processing, it will significantly degrade the efficiency of scene graph generation.
To overcome these limitations, %of multi-stage modeling and simplify the modeling process, 
we aim at the \emph{single-stage} paradigm that properly utilizes both 2D images and 3D point cloud modalities to generate OR scene graphs.

Numerous generic SGG methods have demonstrated significant success in general computer vision tasks, \eg, graph generation of natural scenes. These methods primarily consist of two main streams: 2D and 3D SGG models. However, both of them exhibit shortcomings when being directly adapted to the surgical domain.
%Although there are numerous end-to-end SGG methods based on deep learning solutions, including 2D and 3D SGG frameworks, none of them can be directly used to be directly applicable to generate scene graphs for OR scenarios. 
The OR scene is more complicated, composed of entities that vary significantly in size and shape—for instance, surgeons compared to anesthesia equipment and surgical instruments. Moreover, these entities are frequently occluded due to the common presence of multiple medical staff members who typically surround the patient throughout the surgical procedure.
The existing 2D SGG models~\cite{li2021bipartite,lin2020gps,tang2020unbiased,dhingra2021bgt,li2022sgtr,cong2023reltr} though can precept natural scenes accurately, for complex surgical procedures, these SGG models lack geometric information reinforcement to provide a comprehensive perception of the associations between people and medical devices.
%\eg, Visual Genome~\cite{krishna2017visual} and Open Images-V6~\cite{kuznetsova2020open},

On the other hand, typical 3D SGG approaches~\cite{wald2020learning,zhang2021knowledge,wang2023vl} on point cloud data assume the availability of 3D instance labels as known inputs. This assumption significantly increases annotation efforts and is often impractical in a clinical setting.
Moreover, surgical actions performed by the chief surgeon are very subtle, making it difficult for 3D SGG methods to precisely capture specific operations without sufficient semantic information.
%The introduction of 3D dense detections in 3D SGG models further brings additional computational costs. 
% Meanwhile, the computational cost of 3D SGG models is relatively high. 
% In this regard, how to efficiently leverage geometric information and productively dig out 2D semantic cues is crucial for accurate scene graph generation in the OR. 
\blue{In this regard, efficiently leveraging geometric information and effectively extracting 2D semantic cues are crucial for accurate scene graph generation in ORs.} 
%Scene understanding in the operating theatre is a more challenging task compared to scene graph generation of natural scenes. It not only requires efficient utilization of geometric information, but also productive mining of both low- and high-level 2D semantic cues to capture imperceptible surgeon-patient interactions. Therefore, the key issue lies in how to leverage multi-view 2D images and 3D point cloud modalities to directly generate OR scene graphs.
% Concentrating on this issue, we design an end-to-end SGG model for OR scene graphs in our work.
Concentrating on this issue, we strive to design a concise multimodal SGG paradigm for OR scenarios.

In this paper, we present a novel single-stage multi-view bi-modal transformer-based SGG model, termed \ourmodel, for efficient scene graph generation in OR recordings. 
As illustrated in \figref{struc_com}(b), our single-stage end-to-end SGG paradigm enables the efficient generation of scene graphs, providing a more concise and effective solution for scene understanding in surgery. 
In contrast to multi-stage OR-SGG methods~\cite{ozsoy20224d,ozsoy2023labrad}, our model requires no integration of intermediate predicted instances and has no reliance on the supervision of pose estimation labels.
Concretely, we propose a relation-sensitive transformer in the relation decoder of \ourmodel. 
This structure first couples the output embeddings produced by the transformer decoder into a set of subject-object pairs and then incorporates relational trait priors to generate dynamic relation queries. 
By interacting with multi-view memories through cross-attention, our model predicts relationships within subject-object pairs and achieves the final scene graph. 
To better converge and exploit features from different modalities, a geometry-visual cohesion (GVC) module is introduced to assemble 2D multi-view semantic cues and 3D geometric features to generate appearance-structure representations, instead of the general projection operation in most existing works.
In the multi-view unit, we also present a view-sync transfusion (VST) scheme that complements the local fine-level information of multiple views to the main view features to enhance the holistic visual semantic representation and mitigate occlusions.
% We extensively evaluate the SGG results of our \ourmodel~on the 4D-OR dataset. Compared to existing multi-stage models, the proposed single-stage model demonstrates competitive performance.
%, achieving 0.90, 0.88, and 0.89 on the precision, recall, and Macro F1, respectively. 
% Additionally, our model also exhibits significant advantages compared to the generic single-stage SGG models.
% Extensive experimental findings demonstrate that the proposed single-stage model achieves competitive performance.
Extensive experimental findings demonstrate the effectiveness of `single stage'.

Our main contributions are summarized as follows:
\begin{itemize}
% \item We proposed \ourmodel, the first single-stage bi-modal framework for 4D SGG in the operating rooms. It is an effective multimodal model that can directly generate accurate scene graphs with end-to-end training.

\item To our best knowledge, \ourmodel~is the first \emph{bi-modal single-stage framework} for SGG in the operating rooms. The proposed OR-SGG model can directly generate accurate scene graphs with end-to-end training.

\item We develop a geometry-visual cohesion (GVC) operation, which integrates 2D synergic semantic features and 3D point cloud geometric features to unified visual-structure representations. Besides, a view-sync transfusion (VST) scheme is designed to consolidate multi-view visual appearance information. 

\item A relation-sensitive transformer is introduced to perceive relations between subject-object pairs and directly generate scene graphs. Wherein, we propose dynamic relation queries using trait priors and then interacting with multi-view features to exploit potential associations.

\item Extensive validations demonstrate that our \ourmodel~achieves superior performance on the 4D-OR benchmark, whether compared to multi-stage OR-SGG methods or to generic single-stage SGG models.
\end{itemize}

\section{Related Work}

\subsection{Semantic Scene Graph in OR}
The main literature in surgical data science focuses on analyzing the surgical scene, such as surgical workflow and tool recognition~\cite{czempiel2020tecno}, instrument segmentation~\cite{colleoni2021robotic}, and scene understanding~\cite{jin2022exploring}. 
Scene understanding in dynamic operating room (OR) environments is a new challenging task, which builds entity interactive systems by semantically modeling the holistic space~\cite{ozsoy20224d,ozsoy2023labrad}.
Özsoy \etal~\cite{ozsoy20224d} first revealed the challenge of SSG generation for the OR scene.
Specifically, they built a new scene graph dataset (4D-OR) for knee surgery scenarios, which consists of multi-view RGB frames and point cloud data accompanied by human pose labels, 3D object bounding boxes, and scene graph labels.
Accordingly, they also proposed a multi-stage SGG baseline, which first predicts the human pose and object proposals by VoxelPose~\cite{tu2020voxelpose} and Group-Free~\cite{liu2021group} respectively, and then combines the prediction labels and feeds them into the 3DSSG model~\cite{wald2020learning} for scene graph prediction.
Building upon the 4D-OR baseline, LABRAD-OR~\cite{ozsoy2023labrad} is a novel lightweight SGG model that introduces the concept of memory scene graphs. This model fuses temporal cues with visual features to improve the SGG accuracy.
% However, both approaches are based on multi-stage architectures, while real clinical scenarios require more on the efficient performance that these models lack, instead of their multi-stage outputs. 
% % it is usually desirable to directly generate the OR scene graph without intermediate steps. 
% % To reach this goal, 
% To this end, we propose a single-stage bi-modal SGG framework. As depicted in \figref{struc_com}, our model takes multi-view inputs and point cloud data to directly generate scene graphs without intermediate processing, enhancing the practicality of OR-SGG models in real-world surgery.
\blue{However, both approaches are based on multi-stage architectures, while in real clinical scenarios, there is a greater demand for streamlined and easily deployable end-to-end multimodal models to simplify the modeling workflow and enhance the flexibility of applications in diverse clinical devices. 
To this end, we propose a single-stage bi-modal SGG framework. As depicted in \figref{struc_com}(b), our model takes multi-view inputs and point cloud data to directly generate scene graphs without intermediate processing, enhancing the practicability of OR-SGG models in real-world surgery.}

\begin{figure*}[t!]
\centering
\includegraphics[width=\linewidth]{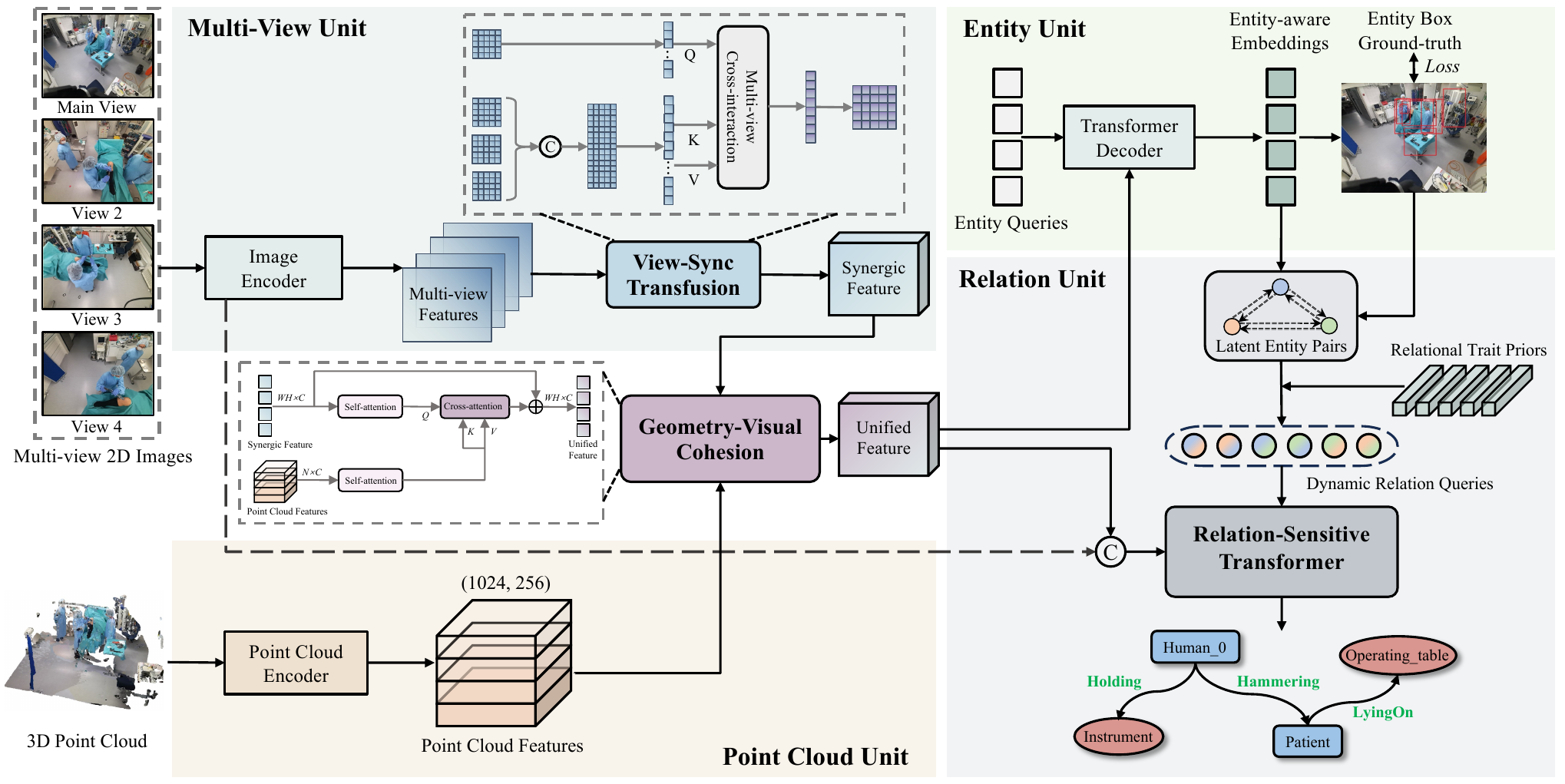}
\caption{Overview of the proposed single-stage multi-view bi-modal \ourmodel~for scene graph generation from operating rooms. We first extract appearance and geometric features separately based on 2D multi-view images and 3D point cloud inputs. In the multi-view unit, a \textbf{View-Sync Transfusion (VST)} is suggested for synthesizing multi-view semantic features. Then, we introduce \textbf{Geometry-Visual Cohesion (GVC)} to fuse 2D synergic features and 3D point cloud features to obtain unified features. In the entity unit, we utilize the unified features and entity queries to predict entity proposals; in the relation unit, we generate \textbf{dynamic relation queries} by assembling latent entity pairs with relational trait priors, which are fed into our \textbf{relation-sensitive transformer} to generate scene graphs in the operating theatre.}
\label{framework}
\end{figure*}
% 给定2D multi-view images和3D point cloud输入，我们首先通过multi-view unit和point cloud unit来分别提取semantic和spatial结构特征。在multi-view unit中，View-Sync Transfusion (VST)被提出用于合成multi-view语义特征。Then, we present a Geometry-Visual Cohesion (GVC)来融合2D synergic特征和3D点云特征来得到unified特征。In the entity unit, 我们借助unified特征和entity查询来预测entity proposals。In the relation unit，我们通过组合latent entity pairs with relational trait priors来生成dynamic relation queries，然后将其输入到我们的relation-sensitive transformer中生成手术室中的场景图。

\subsection{2D Scene Graph Generation}

The main purpose of Scene Graph Generation (SGG) is to construct representations of visual relationships between objects, which has been extensively studied in the field of computer vision in 2D images~\cite{chang2021comprehensive}. 
In the deep learning era, mainstream SGG approaches~\cite{tang2020unbiased,li2021bipartite,lin2020gps,dhingra2021bgt} primarily adopt a two-stage framework, firstly locating objects using an object detection network, and then employing various mechanisms to purify contextual structural information and refine location-specific features, \eg, message passing~\cite{li2021bipartite}, attention module~\cite{dhingra2021bgt, lu2021context}, and visual translation~\cite{zhang2017visual}, to predict relationships by classifiers and generate scene graphs.
Recently, inspired by fully convolutional network (FCN) in object detection models~\cite{long2015fully}, one-stage SGG methods employ FCN-based~\cite{liu2021fully,ozsoy2023location} or CNN-Transformer architectures~\cite{li2022sgtr,cong2023reltr} to efficiently predict visual relationships directly from global features using sparse proposal sets and queries, making this paradigm concise and easy to converge.
% For instance, Liu \etal~\cite{liu2021fully} proposed a fully convolutional SGG approach that applies relation affinity fields to encode both semantic and geometric features towards simultaneous detection of objects and relations.
\blue{Subsequently, Teng \etal~\cite{sparsercnn} proposed a structured sparse R-CNN framework based on set prediction to directly generate scene graphs without the need for explicit object detection and preceding graph construction.} 
To further improve performance, subsequent one-stage models, such as SGTR~\cite{li2022sgtr} and RelTR~\cite{cong2023reltr}, utilize the Transformer architecture and the attention mechanism to capture global perception and predict triplets by interacting with subject-object queries.
\blue{Furthermore, Pix2SG~\cite{ozsoy2023location} is proposed to allow the model to make multiple prediction sequences sequentially from a set of objects like an autoregressive decoder and leverages a transformer-based sequence generation design to liberate the dependence on spatial localization.}
However, 2D images lack critical geometric information for SGG in ORs.
For better capturing location-related relationships, our model incorporates 3D point cloud inputs that complement 2D semantic information from the images to enhance the perception of structural correlations.

\subsection{3D Scene Graph Generation in Point Cloud}

In contrast to 2D SGG at the image level, feeding point clouds to 3D SGG models provides a more robust representation of the geometric properties of objects~\cite{wang2023vl}, leading to a higher sensitivity in predicting structure-aware relationships. 
Wald \etal~\cite{wald2020learning} developed a large-scale 3D semantic scene graph (SSG) dataset, termed 3DSSG. They also proposed an end-to-end SGG pipeline for constructing 3D scene graphs via graph convolution networks (GCNs). 
Later, Zhang \etal~\cite{zhang2021knowledge} introduced an approach for directly generating scene graphs from 3D point clouds, which encodes the point clouds into more discriminative latent spaces as a priori knowledge. Nevertheless, it is challenging to learn semantic information from the point cloud, since the point cloud is defined on geometric structures instead of more semantic colors or textures. 
% However, relying solely on the point cloud modality captures limited geometric structures without fully exploiting semantic features. 
For this reason, VL-SAT~\cite{wang2023vl} incorporates 2D instance labels for multimodal training, which assists the 3D-only SGG model in capturing richer semantic information.
It is worth noting that the 3DSSG benchmark generally assumes the instance labels as known inputs during model training.
In contrast, for the 4D-OR benchmark, instance labeling in the operating theatre is difficult and laborious to acquire.
% This drives us to design a bi-modal SGG model that takes both 2D images and 3D point clouds as inputs and directly produces a comprehensive scene understanding.
This drives us to design a bi-modal SGG model that complements 3D modalities by 2D semantics to generate a comprehensive scene understanding without the need for known instance labels.

\section{Methodology}

The overall architecture of the proposed single-stage bi-modal \ourmodel~is described in \figref{framework}. For the input of four view images, we employ a CNN-Transformer encoder to extract multi-view semantic features.
Subsequently, we introduce a view-sync transfusion (VST) scheme that complements the multi-view information into the main view to attain comprehensive and enriched synergic features.
Bilaterally, for 3D point cloud input, we adopt PointNet++~\cite{qi2017pointnet++} as the point encoder to capture abundant geometric features. 
Given 2D and 3D modal representations, we design a geometry-visual cohesion (GVC) operation to implicitly interact synergic semantic information with point cloud features to achieve unified appearance-structure features. 
% After that, we leverage the unified feature in the entity and relation unit to yield entity proposals and predict relationships. 
After that, we leverage the unified feature to yield entity (\ie, subject and object) proposals and predict relationships. In the operating room, the entities include surgeons, patients, operating tables, instruments, etc.
In the entity unit, we initialize a set of entity queries and produce entity proposals with the instance decoder in line with DETR~\cite{carion2020end}. 
In order to predict relationships without relying on predictions from intermediate stages, we propose a relation-sensitive transformer decoder to perceive interactions between entity pairs and generate OR scene graphs. 
We will detail the structure of VST, GVC, and relation-sensitive transformers embedded in \ourmodel~in the following subsections.

\subsection{Multi-View and bi-modal Interaction}

\ourmodel~comprises two encoder units to process multi-view 2D images and 3D point cloud data of the whole scene, respectively.
To utilize the multi-view representation to comprehensively emphasize surgical operations and enhance structure perception through holistic point cloud samples, we introduce a view-sync transfusion (VST) scheme to synergize multi-view features and a geometry-visual cohesion (GVC) operation to complement the geometric information.

\subsubsection{Multi-View Unit with VST}

In the multi-view image encoder, we utilize four different camera viewpoints \{$I_{i}$\}$_{i=1}^{4}$ as the 2D inputs. 
The multi-view images are initially passed through ResNet-50~\cite{he2016deep} to extract semantic features.
Following the encoder design of DETR~\cite{carion2020end}, we flatten the last layer features $R_{i}^{5}$ from ResNet-50 for each viewpoint. We leverage a transformer encoder to enhance the global perception of different views.
The transformer encoder consists of six layers, containing a multi-head self-attention module and a feed-forward network (FFN) in each layer. 
% After the shape-restoring operation, we can produce \textcolor{red}{high-level semantic features \{$F_{e}^{i}$\}$_{i=1}^{4}$$\in \mathbb{R}^{w\times h\times 256}$ for all four viewpoints}.
After the shape-restoring operation, we can produce high-level semantic features for all four viewpoints.
While \emph{View}\#1 provides a broader receptive field, several critical surgical operations (\eg, hammering, drilling, cutting) require specific viewpoints to focus on localized situations.
Hence, it is essential to synergize features from more viewpoints for a comprehensive understanding of the whole OR. 

With this in mind, we propose a VST scheme in the multi-view unit to integrate multi-view features extracted from the transformer encoder, generating the synergic feature $F_{s}$.
Among features from the four viewpoints, \emph{View}\#1 embraces a more expansive scene, thus we consider it as \emph{Main View} while the other views (\ie, \emph{View}\#2, \emph{View}\#3, and \emph{View}\#4) serve as auxiliary viewpoints to complement the main view.
Inspired by the cross-attention operation in transformers~\cite{vaswani2017attention,chen2021mvt}, we introduce a multi-view cross-interaction module to integrate shared context information between the main and auxiliary views.
%As described in \figref{framework}, 
Specifically, we treat the flattened main-view feature $F_{m}\in \mathbb{R}^{wh\times 256}$ as the Query. 
% In counterpart, the features of \emph{View}\#2, \emph{View}\#3, and \emph{View}\#4 ($F_{a}^{j}, j=2,3,4$) are concatenated and flattened to form an auxiliary viewpoint vector $F_{\alpha}\in \mathbb{R}^{3wh\times 256}$, serving as the Key and Value $F_{\alpha}=[F_{a}^{2},...,F_{a}^{j}]\in \mathbb{R}^{3wh\times 256}$.
\blue{In counterpart, the features of \emph{View}\#2, \emph{View}\#3, and \emph{View}\#4 ($F_{a}^{j}, j=2,3,4$) are concatenated and flattened to form an auxiliary viewpoint vector $F_{\alpha}=[F_{a}^{2},...,F_{a}^{j}]\in \mathbb{R}^{3wh\times 256}$, serving as the Key and Value.}
% where $[\cdot]$ is the concatenation operation.
Through the multi-view cross-interaction operation, we implement the implicit interaction between global features from the main view and other auxiliary views.
This operation can be formulated as
\begin{equation}
\mathcal{A}(F_{m},F_{\alpha})=Softmax(\frac{W_{q}F_{m}(W_{k}F_{\alpha})^{T}}{\sqrt{d_{k}}})W_{v}F_{\alpha},
\end{equation}
where $W_{q}$, $W_{k}$, and $W_{v}$ represent linear mapping matrices and $d_{k}$ is the dimension of $F_{m}$ after linear transformation. 
Then, the aggregated features are performed with the main view features for a residual connection and layer normalization to enhance holistic information, followed by an FFN and layer normalization to generate the synergic feature $F_{s}\in \mathbb{R}^{wh\times 256}$.

\subsubsection{Point Cloud Unit with GVC}

In parallel, to enrich the geometric information available for improving the structural relationship prediction, we add 3D point cloud inputs to our point encoder unit.
% For an input point cloud $P\in \mathbb{R}^{N\times 3}$ with $N$ points, we employ PointNet++~\cite{qi2017pointnet++} as the 3D backbone to encode each point spatially and downsample to obtain the point feature $F_{p}\in \mathbb{R}^{1024\times 256}$ with abundant geometric information.
For an input point cloud $P\in \mathbb{R}^{N\times 3}$ with $N$ points, we employ PointNet++~\cite{qi2017pointnet++} as the 3D backbone to sample and group the input point cloud to a feature space of 256 dimension and 1024 points, resulting in the point cloud feature $F_{p}\in \mathbb{R}^{1024\times 256}$.

After obtaining 2D synergic features and 3D point cloud features, we can generate complementary information from both modalities for sufficient integration of appearance and geometric cues.
% Previous works~\cite{ozsoy20224d,ozsoy2023labrad} utilize 2D and 3D samples for pose estimation and object detection, respectively, and then combine predicted instance labels.
Previous works~\cite{ozsoy20224d,ozsoy2023labrad} utilize 2D and 3D samples for pose estimation and object detection, respectively, and then combine the predicted instance labels of humans and objects to generate scene graphs via GCN~\cite{wald2020learning}.
To enable our single-stage framework for training efficiency while leveraging 3D representation, 
%To ensure the training efficiency of our single-stage framework while leveraging 3D information, 
we inject point cloud features into the synergic feature to facilitate supervision by only utilizing image-level ground truth.
In this regard, we design the GVC operation to implicitly heterogeneously cohesion the fused multi-view synergic features with the 3D point cloud features rather than the direct projection manipulation.
%As shown in~\figref{framework}, 
Specifically, the synergic 2D feature $F_{s}$ and the point cloud feature $F_{p}$ are aligned in the self-attention module at different feature levels, respectively. 
Subsequently, the cross-attention operation is employed to interact and blend features from different modalities, where the Key and Value are both point cloud feature vectors and the Query is the synergic 2D feature vector.
Finally, a residual connection is applied to consolidate semantic information. The entire process of GVC can be described as follows:
\begin{equation}
F_{u}=F_{s}+\mathcal{CA}(\mathcal{SA}(F_{s}),\mathcal{SA}(F_{p})),
\end{equation}
where $\mathcal{SA}$ and $\mathcal{CA}$ stand for self-attention and cross-attention operations, respectively. The unified feature $F_{u}$ comprises sufficient 2D semantic cues and 3D structural information for predicting entities and generating scene graphs.

\subsection{Entity Pair Generation}

\subsubsection{Entity Predictions}

Before predicting relationship triplets [entity$_{sub}$, relation, entity$_{obj}$] which contain the \emph{relation} between \emph{subject entity} and \emph{object entity} in the operating theatre, we first generate entity boxes in the entity unit, allowing for the subsequent pairing of entities to predict the relationships in the surgical scene.
%we first generate a set of entity-aware embeddings as well as entity proposals in the entity unit, allowing for the subsequent pairing of entities to predict the relationships in the surgical scene.
% We utilize a set of randomly initialized queries and the transformer decoder to generate entity-aware embeddings.
Following DETR~\cite{carion2020end}, we utilize a set of randomly initialized queries as input to the transformer decoder.
% As described in \figref{framework}, we exploit the unified feature $F_{u}$ in the transformer decoder to interact with queries and optimize the entity proposal reasoning.
In the transformer decoder, we interact the unified feature $F_{u}$ with queries to produce the output embedding via cross-attention operations. Then an FFN is used to get entity proposals with class labels.
During the training phase, we supervise the prediction of entity boxes by 2D box labels.
% It is worth noting that box supervision is trained together with our model without separate fine-tuning.
It is worth noting that box supervision is trained together with the final scene graph labels without additional pre-training.

\subsubsection{Dynamic Relation Queries}

\begin{figure}[t!]
\centering
\includegraphics[width=\linewidth]{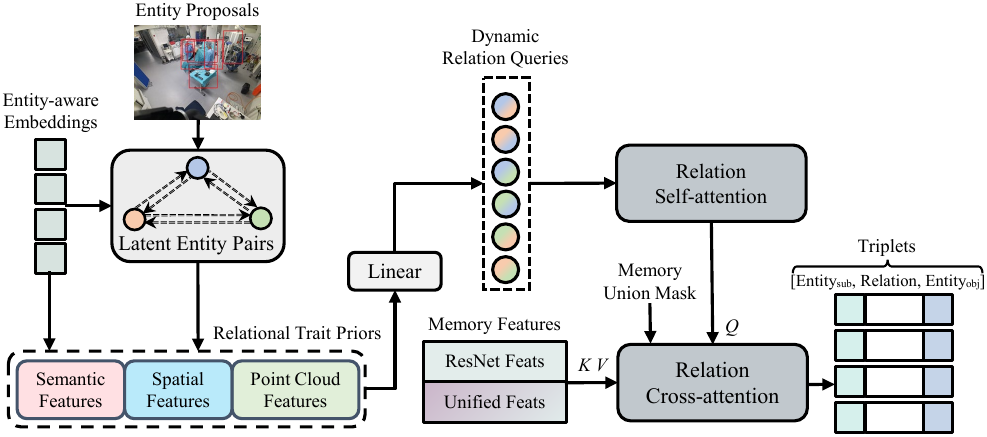}
\caption{Illustration of our relation-sensitive transformer decoder.}
\label{stru_rel}
\end{figure}

The previous multi-stage OR-SGG models~\cite{ozsoy20224d,ozsoy2023labrad} first predict different types of instance labels (\ie, human pose estimation labels and object detection labels), and then merge them through intermediate steps before feeding them into the subsequent SGG network.
% After obtaining entity proposals, we directly build entity connections to predict potential relationships between entity pairs. 
% without computing instance labels.
In contrast, our framework proposes to establish entity connections after generating entity proposals and then predicts potential relationships between entity pairs.
Inspired by the concept of building edges between nodes in graph convolutional networks (GCN), in relation unit, we construct all potential subject-object pairs.
For the subject and object in each pair, we extract semantic features, spatial features, and supplemental point cloud features within the entity regions to generate relational trait priors and then yield dynamic relation queries (see~\figref{stru_rel}).
Specifically, we first filter out $N$ entities with class scores greater than 0.5 and pair them to generate $N(N-1)$ entity pairs.
Here the number of $N$ is not fixed and is dynamically changed according to the perception of entity proposals in the entity unit.
To obtain corresponding semantic features, we concatenate the entity-aware embeddings associated with entities.
Spatial features are calculated based on the positional attributes between entities, including differences in centroid coordinates, distances between the coordinates of center points, and bounding box areas. 
Given a pair of entities with center point coordinates $(E_{x}^{i}, E_{y}^{i})$ and $(E_{x}^{j}, E_{y}^{j})$, the spatial feature $S_{ij}$ can be computed as
\begin{equation}
\bar{x}=E_{x}^{i}-E_{x}^{j}, \quad\bar{y}=E_{y}^{i}-E_{y}^{j},
\end{equation}
\begin{equation}
S_{ij}=[\bar{x}, \bar{y}, \sqrt{\bar{x}^{2}+\bar{y}^{2}}), Ai, Aj],
\end{equation}
where $Ai$ and $Aj$ are areas of two entity boxes, respectively.
Furthermore, we also embed the supplemental point cloud features to enhance the structural representation.
We then concatenate the three kinds of entity features and use a multi-layer perception (MLP) to obtain relational trait priors:
\begin{equation}
V_{ij}=MLP([R_{ij}, S_{ij}, P_{ij}]) \in \mathbb{R}^{1\times 256},
\end{equation}
\begin{equation}
Q_{r}=[V_{1},...,V_{N(N-1)}] \in \mathbb{R}^{N(N-1)\times 256},
\end{equation}
where $R_{ij}$ denotes the semantic feature, $P_{ij}$ is the point cloud feature, and $V_{ij}$ indicates the relational trait prior for per entity pair.
Then, we concatenate the relational trait priors of all entity pairs to generate dynamic relational queries $Q_{r}$.

\subsection{Relation-Sensitive Transformer}\label{RST}

Unlike previous works~\cite{qi2019attentive,ozsoy20224d} that predict relationships via GCN, we input dynamic relation queries into the relation-sensitive transformer block to predict predicates between subject-object pairs.
As illustrated in~\figref{stru_rel}, the dynamic relational query $Q_{r}$ is first fed into the self-attention module to enhance the relational representation of each triplet, obtaining intermediate features $Q_{r}^{'}$.
Considering that relation queries are insensitive to location information during self-attention operation, we do not input position encoding~\cite{vaswani2017attention} into the transformer block. 
After that, we utilize cross-attention operation to interact $Q_{r}^{'}$ with 2D semantic features.
% Here, we use $Q_{r}^{'}$ as the Query and the unified feature $F_{u}$ as the memory feature. 
Here, we use $Q_{r}^{'}$ as the Query and the unified feature $F_{u}$ is regarded as the encoder memory feature for the Key and Value.
%For more concentration on the relationship between the surgeon and the patient, we also incorporate features from \emph{View}\#4, into the memory feature to optimize the prediction of some action relationships (\eg, touching, suturing, and cementing).
Since (\emph{View}\#4) concentration on the operation relationship between the surgeon and the patient, we also incorporate (\emph{View}\#4) features into the memory features to optimize predictions of some action relationships (\eg, touching, suturing, and cementing).
As shown in \figref{framework}, we concatenate the \emph{View}\#4 features extracted from the last layer of ResNet with the unified feature: $\mathcal{M}$=$[F_{u}, R_{4}] \in\mathbb{R}^{2wh\times 256}$.
\blue{Combining CNN (Convolution Neural Network) features essentially serves to supplement encoder memory features with lower-level features.}
The memory feature $\mathcal{M}$ is used as the Key and Value.
\secref{ablation} discusses the effectiveness of this operation in detail with the ablation study.
Additionally, to improve the efficiency of cross-attention, inspired by masked attention from Mask2Former~\cite{cheng2022masked}, we retain only the union of foreground regions of entities for the attention operation. The relation cross-attention operation can be expressed as
\begin{equation}
\mathcal{A}(Q_{r}^{'},\mathcal{M})=Softmax(\frac{W_{q}Q_{r}^{'}(W_{k}\mathcal{M})^{T}+\mathcal{Z}}{\sqrt{d_{k}}})W_{v}\mathcal{M},
\end{equation}
\begin{equation}
\mathcal{Z}=\begin{cases}
0, & \text{if}\quad(x, y)\in U \\ 
-\infty, & \text{otherwise}
\end{cases},
\end{equation}
where $\mathcal{Z}$ indicates the attention mask and $U$ denotes the union of foreground regions.
$(x, y)$ represents the pixel coordinates.
Following a residual connection, layer normalization, and an FFN, we can obtain the predicted triples and combine them to generate a scene graph.

\subsection{Objective Function}

We train our model in an end-to-end manner. The total objective function of \ourmodel~consists of the entity box loss $L_{ent}$ and the relation classification loss $L_{rel}$ in the relationship-sensitive decoder:
\begin{equation}
\mathcal{L}_{total}=\lambda  \mathcal{L}_{ent}+ \mathcal{L}_{rel},
\end{equation}
where $\lambda$ is the weighting factor. \blue{In line with DETR, $L_{ent}$ contains L1 loss and GIoU (Generalized Intersection over Union) loss~\cite{rezatofighi2019generalized} for box regression and Focal loss~\cite{lin2017focal} ($L_{foc}$) for entity classification.} Here, $\lambda$ is set to 1. In addition, we add the auxiliary loss~\cite{al2019character} after each decoder layer. The relational classification loss is mainly implemented by Focal loss:
\begin{equation}
\mathcal{L}_{rel}=\frac{1}{\sum_{i=1}^{C}g_{i}}\sum_{i=1}^{C}g_{i}L_{foc}(\hat{g_{i}}, g_{i}),
\end{equation}
where $C$ is the number of relation classes. $\hat{{g}_{i}}$ is the predicted probability of the $i$-th relation, and $g_{i}\in \{0,1\}$ indicates whether the label includes the relation category or not.

\section{Experiments}

\subsection{Dataset and Evaluation Metrics}

\subsubsection{Dataset}
We used a large-scale public benchmark dataset 4D-OR~\cite{ozsoy20224d} for method validation.
As a new yet essential task of OR-SGG, to our best knowledge, only 4D-OR dataset~\cite{ozsoy20224d} is currently publicly available yet well-known to be widely used for validation. 
This dataset consists of ten knee replacement surgery procedures with a total of 6,734 scenes. Each scene provides 2D multi-view images from different viewpoints and a 3D point cloud description. 
The ten surgical workflows are divided into three sets, with six takes (4,024 scenes) used for training, two takes (1,332 scenes) used for validation, and two takes (1,378 scenes) used for testing\footnote{The test set is not publicly available and requires using the online addresses \url{https://bit.ly/4D-OR_evaluator} to evaluate SGG results}. 
% The 4D-OR dataset contains rich annotations, including 3D entity boxes for 12 categories in the operating room, scene graph relationship labels for 14 categories, and 6D human pose annotations.
The 4D-OR dataset contains rich annotations, including scene entities with 12 categories in the operating room and relationship labels with 14 categories.
In addition, they also provide 3D entity boxes and 6D human pose annotations for auxiliary supervision.

Since the entity box labels are utilized in 2D format for our single-stage training, in our experiments, we project the 3D box labels annotated in 4D-OR into 2D boxes via camera parameters. 
Similar to the 4D-OR method~\cite{ozsoy20224d}, to compensate for the recognition of medical instruments, we locate the hands of the head surgeon by the left and right wrist coordinates from pose annotations. 
Then, we mark boxes with 100$\times$100 size to produce the instrument annotations.
The 2D supervision labels generated from ready-made annotations will be available at \url{https://github.com/PJLallen/S2Former-OR}.

\subsubsection{Evaluation Metrics}

Consistent with the state-of-the-art methods~\cite{ozsoy20224d,ozsoy2023labrad}, we used precision, recall, and Macro F1 metrics to evaluate predicted scene graph relationships. 
We also employed the same evaluation metrics as~\cite{ozsoy20224d,ozsoy2023labrad} in the downstream task of clinical role prediction. 
Moreover, in comparison with the generic SGG models on the 4D-OR benchmark, we adopted several widely used metrics for generic SGG model evaluation, including recall metric under the IoU threshold of 0.5 (R$@$50), weighted mean average precision of relationship detection (wmAP$_{rel}$), and phrase detection (wmAP$_{phr}$) from Open Images Challenge~\cite{kuznetsova2020open} for a more comprehensive evaluation.

\subsection{Implementation Details}

We train the proposed model for 80 epochs on 2 RTX3090 GPUs with an AdamW optimizer. 
The batch size per GPU is set to 2, and the initial learning rate is 5e-5 with a weight decay of 0.0001.
We adopt ResNet-50~\cite{he2016deep} and the DETR encoder~\cite{carion2020end} pretrained on MS-COCO as the multi-view image encoder.
The point cloud encoder and other components are randomly initialized.
During training, the short edges of the input multi-view images are randomly resized between 480 and 800 with aspect ratio unchanged.
Additionally, we apply color jitter and random horizontal flip for data augmentation.
During validation and inference, the short edges are fixed to 480 with no data augmentation.
In the transformer and relation-sensitive decoders, we use six multi-head attention layers with 256 hidden dimensions.
The number of entity queries is set to 20 and the IoU threshold in NMS is 0.7.

\begin{table*}[!t]
\centering
\footnotesize
\renewcommand{\arraystretch}{1.7}
\renewcommand{\tabcolsep}{0.6mm}
\caption{\blue{Detailed comparisons with the state-of-the-art OR-SGG methods on 4D-OR test set. \emph{Params} and \emph{Avg} stand for parameters and average value, respectively. The best scores for each metric are highlighted in bold.}}
\label{tab:sota1}
      \smallskip\noindent
  \resizebox{\textwidth}{!}{
\begin{tabular}{c|l|c|c|c|cccccccccccccc|c}
\hline
\multicolumn{1}{c|}{Stage}                        & Methods                   & Params & \blue{Time/s} & Metrics  & Assist & Cement & Clean & CloseTo & Cut & Drill & Hammer & Hold & LyingOn & Operate & Prepare & Saw & Suture & Touch & Avg \\ \hline
\multirow{6}{*}{\begin{sideways}Multi-Stage\end{sideways}} & \multirow{3}{*}{4D-OR~\cite{ozsoy20224d}}      & \multirow{3}{*}{84.8M} &  &  Precision     & 0.42   &  0.78  & 0.53   &  \textbf{0.97}   &   0.49  &  0.87   &   0.71   &   0.55  & \textbf{1.00}    &   0.55   & 0.62   &  0.69   &   0.60    &   0.41   & 0.68    \\ 
\multicolumn{1}{c|}{}                             &                             &  & \blue{1.28} & Recall  & \textbf{0.93}   &  0.78   &   0.63  &  0.89   &  0.49   &   \textbf{1.00}   &   0.89    &  \textbf{0.95}    &   0.99   &   \textbf{0.99}    &    \textbf{0.91}  &    0.91  &   \textbf{1.00}  &   0.69   &   0.87    \\ 
\multicolumn{1}{c|}{}                             &                             &   &    &  Macro F1 & 0.58   &  0.78    &    0.57  &   0.93   &  0.49   &  0.93   &   0.79   &   0.70   &   0.99    &   0.71    &   0.74   &  
 0.79   &    0.75    &  0.51     & 0.75    \\ \cline{2-20}                      & \multirow{3}{*}{LABRAD-OR~\cite{ozsoy2023labrad}}  & \multirow{3}{*}{85.8M}  &  &   Precision     & 0.60   &   0.96   &  0.86    &   0.96   & \textbf{0.91}    &    \textbf{1.00}   &    0.93    &  0.71    &  \textbf{1.00}      &    0.85    &    0.77     &   0.78  &   \textbf{1.00}   &   0.71  & 0.87    \\ 
    &         &  & \blue{1.46} &  Recall      & 0.86   &   0.93     &   0.72    &   \textbf{0.94}   &   0.68  &  0.94   &  \textbf{0.95}   &  \textbf{0.95}    &   \textbf{1.00}    &   \textbf{0.99}    &  \textbf{0.91}    &  0.93   &     \textbf{1.00}   &    \textbf{0.72}   & \textbf{0.90}    \\ 
  &        &  &  & Macro F1 & \textbf{0.71}   &   0.94     &   0.78    &    \textbf{0.95}   & 0.78    &   \textbf{0.97}  &  \textbf{0.94}      &   \textbf{0.81}   & \textbf{1.00} &   \textbf{0.91}    &   0.84   &  0.85   &     \textbf{1.00}   &    \textbf{0.71}   & 0.88    \\ \hline
\multirow{3}{*}{\begin{sideways}Single-Stage\end{sideways}}                       
  & \multirow{3}{*}{\textbf{\ourmodel}} &  \multirow{3}{*}{\textbf{60.6M}} &  &  Precision     & \textbf{0.65}   &   \textbf{1.00}    &  \textbf{0.91}  &   \textbf{0.97}    &   0.82  &   0.98    &   \textbf{0.94}    &  \textbf{0.79}   &   \textbf{1.00}    &   \textbf{0.90}    &    \textbf{0.83}   &   \textbf{0.93}  &  0.94    &  \textbf{0.81}     & \textbf{0.90}    \\ 
   &          & & \blue{\textbf{0.98}}  &   Recall      & 0.66   & \textbf{0.94}     &   \textbf{0.86}    &  0.93    &  \textbf{0.85}   &     0.92  &    0.94    &   0.83   &    \textbf{1.00}    &      0.91   &  0.89       &  \textbf{0.99}   &      0.97  &   0.49    & 0.88    \\
&           & &   &  Macro F1  & 0.66   &  \textbf{0.97}      &  \textbf{0.89}     &   \textbf{0.95}    &  \textbf{0.84}   &   0.95    &  \textbf{0.94}  &   \textbf{0.81}   &   \textbf{1.00}      &   \textbf{0.91}      &    \textbf{0.86}   &  \textbf{0.96}   &   0.95   &  0.61   & \textbf{0.89}    \\ \hline
\end{tabular}}
\end{table*}

\subsection{Comparison with State-of-the-Arts}

To evaluate the performance of our \ourmodel~for the OR-SGG task, we carried out detailed comparisons with existing SGG approaches tailored to OR scenes and generic single-stage 2D SGG models on the 4D-OR benchmark. 
For a fair comparison, we uniformly use the official code to train and evaluate each model on 4D-OR validation and test sets.

\begin{figure*}[t!]
\centering
\includegraphics[width=\linewidth]{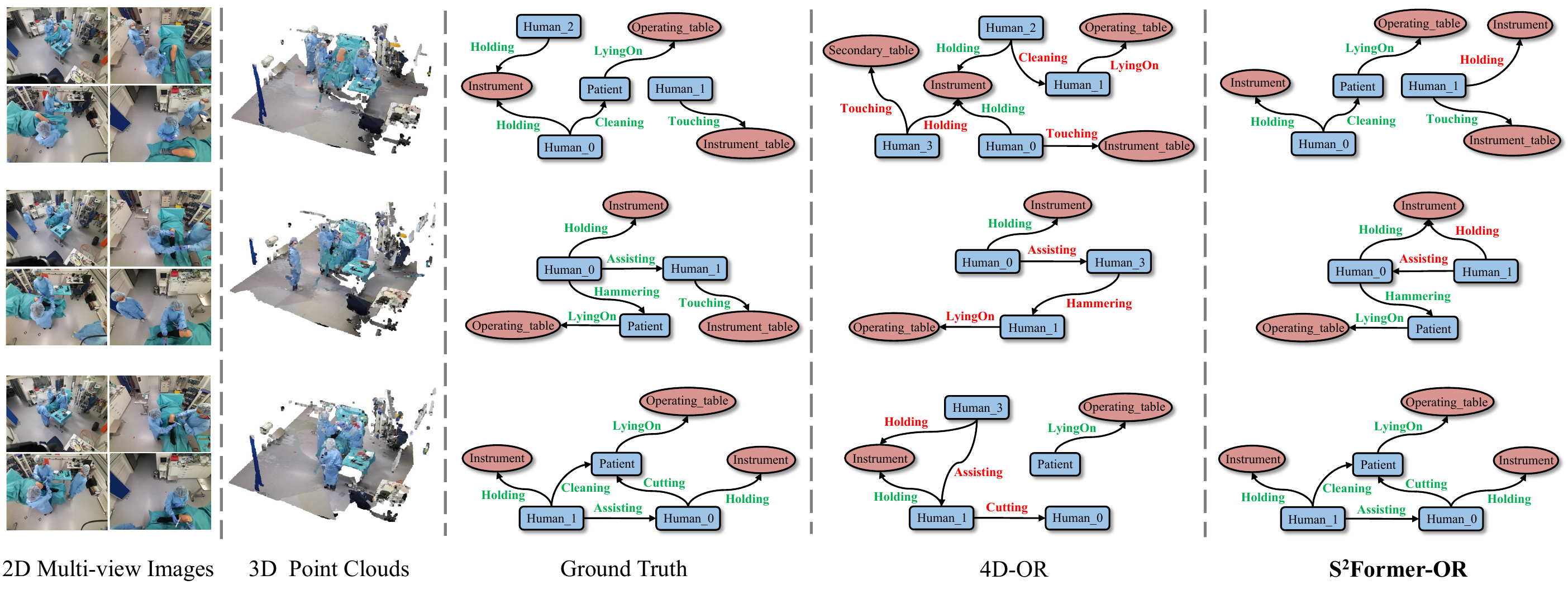}
\caption{\blue{Qualitative results of the 4D-OR model~\cite{ozsoy20224d}, LABRAD-OR~\cite{ozsoy2023labrad}, and our \ourmodel~on the 4D-OR validation set. The blue rectangles represent the human attribute and the red ellipses denote the object attribute. Correct/Erroneous predicted relationships are shown in green/red.}}
\label{visual}
\end{figure*}

\subsubsection{Comparison with OR-SGG Methods}

We conducted a comprehensive comparison between the proposed \ourmodel~and existing OR-SGG models, \ie, 4D-OR~\cite{ozsoy20224d} and LABRAD-OR~\cite{ozsoy2023labrad}.
\tabref{tab:sota1} exhibits quantitative results of our model and competitors in terms of the precision, recall, and macro F1 metrics. 
We can observe that \ourmodel~demonstrates promising performance, particularly in average precision and macro F1 score, achieving 0.90 and 0.89, respectively.
Benefiting from the thorough fusion of semantic features and structural information, our model excelled in predicting critical surgeon-patient actions, \eg, achieving Macro F1 scores of 0.97, 0.91, and 0.96 in cementing, operating, and sawing relations, respectively. 
% More importantly, \ourmodel~acts as a single-stage end-to-end model, enabling the direct generation of scene graphs in practical applications by feeding surgical video samples without intermediate prediction steps.
% More importantly, \ourmodel~acts as an end-to-end model with fewer parameters, enabling the direct generation of scene graphs in practical applications by feeding surgical data without intermediate prediction processes (\eg, no need for pose estimation labels).
% As a single-stage architecture, the inference time of our model is also faster than the 4D-OR model by 0.3s per frame, significantly increasing the potential of model deployment in real clinical practice.
\blue{More importantly, the proposed~\ourmodel, acting as a bi-modal end-to-end model, is capable of directly generating scene graphs by feeding surgical data without intermediate prediction processes (\eg, no need for pose estimation). Furthermore, As illustrated in the third and fourth columns of~\tabref{tab:sota1}, our single-stage model has advantages over multi-stage architectures in terms of parameters and inference time, \eg, 0.3 seconds per frame faster than the 4D-OR model, significantly boosting the potential of model deployment in real clinical practice.}

To further intuitively evaluate the effectiveness of generated scene graphs, we visualize the prediction results of some keyframes in \figref{visual}.
Notably, we omitted the ``CloseTo" relationship to explicitly display the scene graph that is most relevant to the surgical procedure.
We can see that our model consistently predicts the interactions of entities without missing key messages in complex operating room environments.
Compared to state-of-the-art OR-SGG models, our model exhibits more accurate detection of entities, avoiding the influence of erroneous entities in predicting relationships. 
While still producing redundant predictions in some cases, our model can predict the overall structure of the scene graph more precisely and robustly compared with others.
%performs robustly for the overall construction of scene graphs.

\subsubsection{Comparison with Single-Stage SGG Methods}

% To further validate the robustness of our method, we also compare with the cutting-edge single-stage SGG methods in a generic scenario of natural scene generation (\ie, SGTR~\cite{li2022sgtr} and RelTR~\cite{cong2023reltr}). We reimplemented them on the 4D-OR dataset.
%简单介绍SGTR RelTR
% Both of them are novel single-modal end-to-end architectures based on transformers.
% As shown in \tabref{tab:sota2}, our model not only achieves better results in vanilla metrics but also outperforms other competitors in R$@$50, wmAP$_{rel}$, and wmAP$_{phr}$ metrics.
% This can be attributed to the interaction of multi-view features via the proposed VST scheme and the extraction of geometric information from point cloud modalities by our GVC operation.
\blue{To further validate the robustness of our method, we also compare with the cutting-edge single-stage SGG methods in a generic scenario of natural scene generation, including SGTR~\cite{li2022sgtr}, RelTR~\cite{cong2023reltr}, and Structured Sparse R-CNN~\cite{sparsercnn}. 
We retrain and validate these novel single-modal end-to-end architectures on the 4D-OR dataset. 
For a fair comparison, we ablate the values of learning rate and batch size for each model, and other components remain the same as the default settings. As shown in \tabref{tab:sota2}, our model not only achieves better results in vanilla metrics but also outperforms other competitors in R$@$50, wmAP$_{rel}$, and wmAP$_{phr}$ metrics. This can be attributed to the interaction of multi-view features via the proposed VST scheme and the extraction of geometric information from point cloud modalities by our GVC operation.}
Further, the sufficient utilization of relational trait priors in generating dynamic relation queries and the efficient relation-sensitive transformer structure have also brought performance gains to our model.

% \begin{table}
% \begin{center}
% \caption{Quantitative comparisons with generic single-stage SGG methods on the 4D-OR validation set.}
% \label{tab:sota2}
% \footnotesize
% \renewcommand{\arraystretch}{1}
% \renewcommand{\tabcolsep}{0.8mm}
% \begin{tabular}{l|c|ccc|ccc}
% \hline
% Methods  &   Backbones  &  R$@$50    &   wmAP$_{rel}$  & wmAP$_{phr}$   &  Prec   &  Rec   &   F1    \\ \hline
% SGTR~\cite{li2022sgtr}     &       Res101      &  0.57    &   0.51   &  0.69   &  0.69  &  0.70  & 0.69  \\ \hline
% RelTR~\cite{cong2023reltr}     &   Res50     &  0.54   &   0.48  &   0.62   &  0.61  &  0.67  &  0.63  \\ \hline
% \rowcolor[RGB]{222,235,247}
% \textbf{\ourmodel}   &   Res50     &   \textbf{0.71}    &  \textbf{0.63}    &   \textbf{0.83}  &  \textbf{0.90}  & \textbf{0.84}  &  \textbf{0.87} \\  \hline
% \end{tabular}
% \end{center}
% \end{table}

\begin{table}
\begin{center}
\caption{\blue{Quantitative comparisons with generic single-stage SGG methods on the 4D-OR validation set. \emph{Prec}, \emph{Rec}, and \emph{F1} denote the precision, recall, and Macro F1 metrics, respectively.}}
\label{tab:sota2}
\smallskip\smallskip
\footnotesize
\renewcommand{\arraystretch}{1.3}
\renewcommand{\tabcolsep}{0.5mm}
\begin{tabular}{l|c|ccc|ccc}
\hline
Methods  &   Backbones  &  R$@$50    &   wmAP$_{rel}$  & wmAP$_{phr}$   &  Prec   &  Rec   &   F1    \\ \hline
SGTR~\cite{li2022sgtr}     &       ResNet-101      &  0.57    &   0.51   &  0.69   &  0.69  &  0.70  & 0.69  \\ \hline
RelTR~\cite{cong2023reltr}     &   ResNet-50     &  0.54   &   0.48  &   0.62   &  0.61  &  0.67  &  0.63  \\ \hline
SS R-CNN~\cite{sparsercnn}     &   ResNeXt-101     &  0.64   &   0.55  &   0.74   &  0.82  &  0.79  &  0.80  \\ \hline
\rowcolor[RGB]{222,235,247}
\textbf{\ourmodel}   &   ResNet-50     &   \textbf{0.71}    &  \textbf{0.63}    &   \textbf{0.83}  &  \textbf{0.90}  & \textbf{0.84}  &  \textbf{0.87} \\  \hline
\end{tabular}
\end{center}
\end{table}

\subsection{Ablation Study}\label{ablation}

\subsubsection{Effectiveness of Each Component}

\begin{table}[!t]
\begin{center}
\caption{Ablation studies for each component in our \ourmodel~on the 4D-OR validation set, including View-Sync Transfusion (VST), Geometry-Visual Cohesion (GVC), and Relation-sensitive Transformer (RST).}
\label{tab:ablation}
\smallskip\smallskip
\footnotesize
\renewcommand{\arraystretch}{1.2}
\renewcommand{\tabcolsep}{2.8mm}
\begin{tabular}{c|ccc|ccc|c}
\hline
ID  &   VST  &  GVC   &  RST   &  Prec   &  Rec   &   F1  & Params  \\ \hline
1  &         &    &        & 0.71 &  0.60  &  0.65  & \textbf{54.1M} \\ \hline
2  &         &    &    \checkmark    & 0.84 &  0.67 & 0.71  & 54.6M \\ \hline
3  &  \checkmark   &     &   \checkmark  & 0.84 & 0.68  & 0.72  & 57.2M \\ \hline
4  &    &  \checkmark    &   \checkmark  & 0.89 & 0.83  &  0.86 & 58.0M \\ \hline
5  &  \checkmark   &  \checkmark    &   &  0.80  & 0.71  & 0.75  & 60.1M \\ \hline
\rowcolor[RGB]{222,235,247}
6  &  \checkmark   &  \checkmark    &   \checkmark  & \textbf{0.90}  & \textbf{0.84}  &  \textbf{0.87} & 60.6M \\ \hline
\end{tabular}
\end{center}
\end{table}

To validate the effectiveness of each component in our model, we conduct detailed ablation experiments to optimize model performance.
The main components include View-Sync Transfusion (VST), Geometry-Visual Cohesion (GVC), and Relation-sensitive Transformer (RST). 
Note that when ablating the VST module, we replace the input of multi-view images with only the main-view image.
When removing the GVC module, we directly project the extracted point cloud features to synergic features through the intrinsic and extrinsic matrices provided in the dataset.
For our RST part, we adopt a GCN structure from the previous OR-SGG model~\cite{ozsoy20224d} as a replacement to perform relation predictions.
The results in \tabref{tab:ablation} indicate that each proposed component imposes a positive impact on \ourmodel.
Compared to the baseline, our model gains 0.19, 0.24, and 0.22 improvement on the precision, recall, and macro F1 metrics after adding three essential parts (refer to the first and sixth rows).
The third row of~\tabref{tab:ablation} illustrates that the GVC operation improves the generated accuracy by sufficiently integrating point cloud features into 2D semantic features.
Moreover, the fifth row highlights the indispensability of RST, where the dynamic relation queries and relation-aware cross-attention operations make our model more conducive to inferring relations.
Compared to other variants, \ourmodel~has no substantial increase in parameters after embedding all components, while achieving better performance.

% \begin{table}[!t]
% \begin{center}
% \caption{Comparison with different numbers of relation queries on the 4D-OR validation set.}
% \label{tab:num-query}
% \footnotesize
% \renewcommand{\arraystretch}{1}
% \renewcommand{\tabcolsep}{4.5mm}
% \begin{tabular}{c|ccc}
% \hline
% Number  &   Prec  &  Rec  & F1   \\ \hline
% 20  &  0.85 & 0.75  &  0.78  \\ \hline
% 50  &   0.86  & 0.83  & 0.84   \\ \hline
% 100 &  0.87   & 0.83  &  0.85  \\ \hline
% 300 &   0.88  &  \textbf{0.84} &  0.85  \\ \hline
% \rowcolor[RGB]{222,235,247}
% Dynamic  & \textbf{0.90}  & \textbf{0.84}  &  \textbf{0.87}   \\ \hline
% \end{tabular}
% \end{center}
% \end{table}

\begin{table}[!t]
\begin{center}
\caption{\blue{Comparison with different numbers of relation queries on the 4D-OR validation set.}}
\label{tab:num-query}
\smallskip\smallskip
\footnotesize
\renewcommand{\arraystretch}{1.1}
\renewcommand{\tabcolsep}{4.5mm}
\begin{tabular}{c|ccc}
\hline
Number  &   Prec  &  Rec  & F1   \\ \hline
20  &  0.85 & 0.75  &  0.78  \\ \hline
50  &   0.86  & 0.83  & 0.84   \\ \hline
100 &  0.87   & 0.83  &  0.85  \\ \hline
150 &  0.88   & \textbf{0.85}  &  0.86  \\ \hline
200 &  0.89   & 0.84  &  0.86  \\ \hline
300 &   0.88  &  0.84 &  0.85  \\ \hline
\rowcolor[RGB]{222,235,247}
Dynamic  & \textbf{0.90}  & 0.84  &  \textbf{0.87}   \\ \hline
\end{tabular}
\end{center}
\end{table}

\begin{table}[!t]
\begin{center}
\caption{\blue{Comparison with different combinations of viewpoint features in the VST scheme on the 4D-OR validation set. Q, K, and V denote Query, Key, and Value in the operation of VST.}}
\label{moreview}
\smallskip\smallskip
\footnotesize
\renewcommand{\arraystretch}{1.2}
\renewcommand{\tabcolsep}{5.0mm}
\begin{tabular}{l|ccc}
\hline
Q,K,V in VST  &   Prec  &  Rec  & F1   \\ \hline
Q: View\#2 K,V: View\#1,3,4  &  0.82 & 0.77  &  0.79  \\ \hline
Q: View\#3 K,V: View\#1,2,4   &   0.81  & 0.77  & 0.79   \\ \hline
Q: View\#4 K,V: View\#1,2,3  &  0.83   & 0.81  &  0.82  \\ \hline \hline
Q: View\#1 K,V: View\#2   &  0.86 & 0.81  &  0.83  \\ \hline
Q: View\#1 K,V: View\#2,3   &   0.87  & 0.83  & 0.85   \\ \hline
\rowcolor[RGB]{222,235,247}
Q: View\#1 K,V: View\#2,3,4  &  \textbf{0.90}   & \textbf{0.84}  &  \textbf{0.87}  \\ \hline
\end{tabular}
\end{center}
\end{table}

\subsubsection{Number of Relation Queries}

% Relation queries are an essential component in our relation-sensitive transformer.
% To examine the impact of the number of relation queries on relation prediction, we compare the performance of our dynamic relation queries (\ie, $N(N-1)$) with other fixed number queries (\ie, 20, 50, 100, and 300).
% Unlike dynamic queries, to produce a fixed number of queries, we directly pair the entity proposals output in the entity unit without using category scores for suppression.
% Then, we apply an MLP operation to all entity pairs to obtain corresponding confidence values and use the top-K selection to filter out a fixed number of pairs for relation-aware query generation.
% As shown in \tabref{tab:num-query}, with a fixed number of relation queries, the performance of \ourmodel~constantly increases with the number of queries until the number reaches 100.
% Compared to the fixed-amount setting, our dynamic relation queries provide greater adaptability to variable scene interactions, making our relation-sensitive transformer more accurate in graph prediction, \eg, 0.87 vs. 0.85 in macro F1.
\blue{To examine the impact of the number of relation queries on relation prediction, we compare the performance of our dynamic relation queries (\ie, $N(N-1)$) with other fixed number queries (\ie, 20, 50, 100, 150, 200, and 300). 
Unlike dynamic queries, to produce a fixed number of queries, we directly pair the entity proposals output in the entity unit without using category scores for suppression. Then, we apply an MLP operation to all entity pairs to obtain corresponding confidence values and use the top-K selection to filter out a fixed number of pairs for relation-aware query generation. 
As shown in \tabref{tab:num-query}, with a fixed number of relation queries, the performance of \ourmodel~constantly increases with the number of queries until the number reaches 150. For dynamic relation queries, we perform distribution statistics on the number of entities $N$ for each sample from 4D-OR, where the mean value of $N$ is 12.29 and the median value is 13. Thus, the number of dynamic relation queries N*(N-1) is approximately 13$\times$(13-1)=156, which is close to the optimal fixed number of relation queries at 150. 
Compared to the fixed-amount setting, our dynamic relation queries provide greater adaptability to variable scene interactions, making our relation-sensitive transformer more accurate in graph prediction.}

\subsubsection{Different combinations of viewpoints in VST}

\blue{In the proposed VST scheme, we interact the main-view features with the local fine-grained information of other multiple views to enhance the holistic 2D semantic representation and mitigate occlusions. To investigate the effect of different viewpoint features on relationship prediction of~\ourmodel, we conduct an ablation study on combinations of multi-view features in the VST scheme. The experimental results in~\tabref{moreview} show that multi-view information, especially View\#1 and View\#2, can effectively facilitate scene graph generation in ORs. In addition, the selection of the Query, Key, and Value of VST is also critical. Our model performs better when the features of View\#1 are served as Queries and the features from other views are used as Keys and Values. This can be attributed to the main view (View\#1) providing global features that contain various entities and relationships, which is more suitable for feature transfusion as the Query. Other views focus more on specific locations and relationship classes, making them more appropriate as Keys and Values to complement the features that round out the main view.}

\subsubsection{Effect of Relational Trait Priors}

In the relation-sensitive decoder, we inject the latent entity pairs with various subject-object trait priors to accurately express relation queries, including semantic, spatial, and point cloud features.
To validate the effectiveness of each feature representation, we compare the contributions of different combinations of prior trait embeddings on the 4D-OR validation set.
As illustrated in~\tabref{tab:rel-prior}, in addition to the indispensable semantic features, \blue{the spatial feature representation of entity pairs remarkably improves the precision value by 3 percentage points.}
It indicates that spatial features can connect entity pairs with the closest correlations. 
Additionally, point cloud features also contribute to enhancing the structural representation of entity pairs.
By aggregating all relational trait priors, \ourmodel~reaches the best performance.

\begin{table}[!t]
\begin{center}
\caption{Combinations of prior trait embeddings in relation queries.}
\label{tab:rel-prior}
\smallskip\smallskip
\footnotesize
\renewcommand{\arraystretch}{1.2}
\renewcommand{\tabcolsep}{4.0mm}
\begin{tabular}{l|ccc}
\hline
Relational Traits  &   Prec  &  Rec  & F1   \\ \hline
Semantic  &  0.86   &  0.82 &  0.84  \\ \hline
Semantic + Spatial  &  0.89  & 0.83  &  0.86  \\ \hline
\rowcolor[RGB]{222,235,247}
Semantic + Spatial + Point Cloud  & \textbf{0.90}  & \textbf{0.84}  &  \textbf{0.87}   \\ \hline
\end{tabular}
\end{center}
\end{table}

\subsubsection{Memory Feature in Relation-Sensitive Transformer}

\begin{table}[!t]
\begin{center}
\caption{\blue{Ablations for combinations of cross-attention memories in relation-sensitive transformer. $F_{u}$, $R_{2}$, $R_{3}$, and $R_{4}$ represent the unified feature from the GVC operation and  the \emph{View}\#2-\#4 features from ResNet, respectively.}}
\label{tab:memory}
\smallskip\smallskip
\footnotesize
\renewcommand{\arraystretch}{1.2}
\renewcommand{\tabcolsep}{5.8mm}
\begin{tabular}{l|ccc}
\hline
Memories  &   Prec  &  Rec  & F1   \\ \hline
 $F_{u}$   &  0.80   &  0.61   &  0.67  \\ \hline
 \rowcolor[RGB]{222,235,247}
  $F_{u}$ + $R_{4}$    &   \textbf{0.90}    & 0.84    &  \textbf{0.87}  \\ \hline
  $F_{u}$ + $R_{2}$ + $R_{4}$     &  0.89   &  0.85  & 0.86   \\ \hline
  $F_{u}$ + $R_{2}$ + $R_{3}$ + $R_{4}$  &  0.87  &  \textbf{0.87}   &  \textbf{0.87}   \\ \hline
\end{tabular}
\end{center}
\end{table}

Relation queries involve interacting with memory features in the proposed relation-sensitive transformer to perceive relationships between entities in ORs. 
As discussed in \secref{RST}, we also added low-level CNN features from other viewpoints to capture crucial actions.
To investigate the effect of different combinations of cross-attention memories on relationship prediction, we attempt to embed features from additional viewpoints to explore the upper bound of the performance of our relation-sensitive transformer. 
As shown in \tabref{tab:memory}, the combination of the unified feature $F_{u}$ and the \emph{View}\#4 feature $R_{4}$ from ResNet gains better performance.
Memories combined with more view features have not further contributed to the relationship predictions.
% The main reason is that the \emph{View}\#4 information can better complement the main view map. 
% The main reason is that \emph{View}\#4 concentrates on the view of surgeon-patient interaction, and thus better complements the main view of xx.
The main reason is that \emph{View}\#4 concentrates on the view of surgeon-patient interaction, and thus better assisting the main view in complementing the emphasis on the surgical-operative relationship.
Besides, the CNN features also provide lower-level cues to unified features.

\begin{figure}[t!]
\centering
\includegraphics[width=\linewidth]{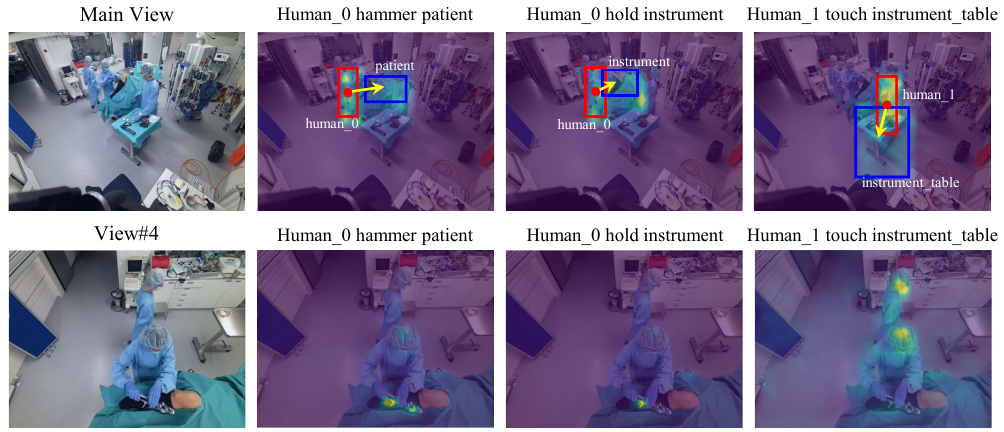}
\caption{Visualizations of attention maps in our relation-sensitive transformer on the 4D-OR validation set. The top row displays the main view weight map with the unified feature $F_{u}$; the bottom row displays the View\#4 weight map with the ResNet feature $R_{4}$.}
\label{attn-map}
\end{figure}

\figref{attn-map} visualizes the attention maps from cross-attention weights in the relation-sensitive transformer.
The top row shows that memory features from the main view can interact comprehensively with entity regions, while the bottom row illustrates that features from \emph{View}\#4 enhance attention toward the surgeon's actions (\eg, hammering and holding).
Specifically, from the third column in \figref{attn-map}, we can see that when the unified features from the main view fail to accurately focus on the instrument region, the \emph{View}\#4 attention map still can capture the position of ``holding" for complementation. 
Whereas, in cases where the unified feature effectively highlights the relationship between entity pairs, assistance from other views is unnecessary.
Overall, combining memory features of $F_{u}$ and $R_{4}$ yields superior performance for relationship predictions.

\begin{table}[!t]
\begin{center}
\caption{\blue{Ablations on different numbers of point cloud features input to the GVC operation on the 4D-OR validation set.}}
\label{morepoint}
\smallskip\smallskip
\footnotesize
\renewcommand{\arraystretch}{1.2}
\renewcommand{\tabcolsep}{4.4mm}
\begin{tabular}{c|ccc|cc}
\hline
Number of Points  &   Prec  &  Rec  & F1 & Time/s   \\ \hline
256 &  0.88   & 0.75  &  0.81   &   \textbf{0.95} \\ \hline
512 &  0.89   & 0.84  &  0.86   &   0.96 \\ \hline
\rowcolor[RGB]{222,235,247}
1024  &   \textbf{0.90}  & 0.84  & \textbf{0.87}  &  0.98 \\ \hline
2048  &  0.89  &  \textbf{0.85}   & \textbf{0.87}  &  1.02 \\ \hline
\end{tabular}
\end{center}
\end{table}

\subsubsection{Number of Point Cloud Features}

\blue{Downsampling to different numbers of point cloud features using PointNet++~\cite{qi2017pointnet++} is crucial for subsequent geometry-visual cohesion (GVC) operation. To this end, we conduct an ablation experiment in~\tabref{morepoint} for different numbers of point cloud features input to the GVC operation, including 256, 512, 1024, and 2048 sampling points. The experimental results demonstrate that as the number of sampling points increases, the contribution of 3D geometric information to the accuracy of scene graph prediction also gradually improves. The performance of our model reaches a plateau when the number of sampling points reaches or exceeds 1024. In order to achieve a better balance between accuracy and computational efficiency, we finally select 1024 points as the default point cloud features for our~\ourmodel.}

\subsection{Clinical Role Prediction}

\begin{figure}[t!]
\centering
\includegraphics[width=\linewidth]{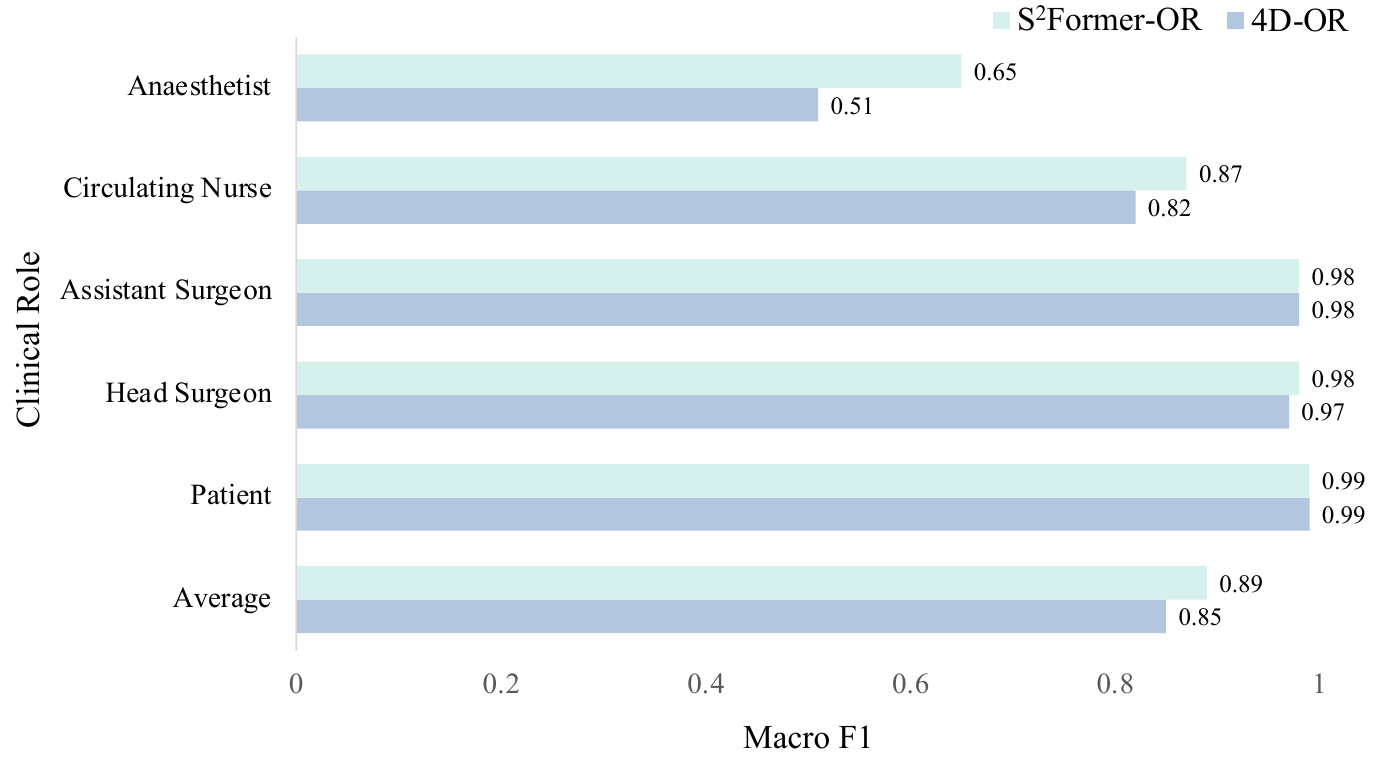}
\caption{Performance comparison of clinical role prediction between our \ourmodel~and the 4D-OR model on the 4D-OR dataset.}
\label{role_comp}
\end{figure}

Assigning role labels to humans in surgical scenes facilitates further improving surgical process monitoring and optimization. 
Since the 4D-OR benchmark provides role labels for each entity associated with the human attribute (\ie, Patient, Head Surgeon, Assistant Surgeon, Circulating Nurse, and Anaesthetist), we perform clinical role prediction based on the generated scene graphs.
For a fair comparison, we also adopt Graphormer~\cite{ying2021transformers} to infer the clinical role probabilities of human entities as in the 4D-OR method~\cite{ozsoy20224d}.
Since our model is a single-stage framework without intermediate results of human pose estimation, we select the predicted entities belonging to the human attribute as human nodes.
\figref{role_comp} compares the performance of our model with 4D-OR on the downstream task of clinical role prediction.
It indicates that \ourmodel~achieves strong performance in all clinical roles, especially in the `Anaesthetist' and `Circulating Nurse' categories, \blue{improving the macro F1 score by 14 and 5 percentage points, respectively.}
Since the anaesthetist is usually located in the corner of operating rooms and appears exclusively at the beginning stage of surgery, there is still room for further improvement.
Overall, our model carries over the advantages of scene graph generation to clinical role prediction.

\subsection{Performance on 3DSSG Benchmark}

\blue{To demonstrate the generalization capabilities of~\ourmodel~and to extend the impact of this work, we also verify the performance of our model on the 3D semantic scene graph (3DSSG) benchmark~\cite{Wald20203ssg} for the 3D indoor scene graph generation. The dataset provides 3D point cloud samples, multi-view RGB inputs, and semantic scene graph annotations.
In this experiment, we use the same data preparation and training/validation partitioning as 3DSSG. The labeling of the 3D instance is known based on the setup of existing related methods~\cite{Wald20203ssg,wang2023vl} in 3DSSG. Thus, we can also project the 3D instance labels into the 2D space by using the camera parameters, thereby obtaining 2D object locations. Then, we directly utilize the position of instances instead of the entity prediction head to generate latent entity pairs. Other settings remain unchanged.
We compare our method with several well-known 3D SGG methods on the 3DSSG validation set, including SGPN~\cite{Wald20203ssg} SGFN~\cite{wu2021scenegraphfusion}, and VL-SAT~\cite{wang2023vl}. For the evaluation metrics, we use the top-k accuracy (A$@$50, A$@$100) and the mean top-k accuracy (mA$@$k) to evaluate the triplet predictions. As shown in~\tabref{tab:3dssg}, our model also achieves competitive results on the indoor 3DSSG dataset, especially on the mA$@$50 and mA$@$100 metrics, demonstrating the strong generalization capability of our~\ourmodel~across general multimodal SGG tasks.}

\begin{table}[!t]
\begin{center}
\caption{\blue{Quantitative comparisons with 3D SGG methods on the 3DSSG~\cite{Wald20203ssg} validation set.}}
\label{tab:3dssg}
\smallskip\smallskip
\footnotesize
\renewcommand{\arraystretch}{1.2}
\renewcommand{\tabcolsep}{3.6mm}
\begin{tabular}{l|cccc}
\hline
Methods   &  A$@$50    &   A$@$100  & mA$@$50   &  mA$@$100    \\ \hline
SGPN~\cite{Wald20203ssg}        &  87.55    &   90.66   &  41.52   &  51.92  \\ \hline
SGFN~\cite{wu2021scenegraphfusion}    &   89.02    &    91.71  &  58.37   &  67.61  \\ \hline
VL-SAT~\cite{wang2023vl}     &  \textbf{90.35}   &   \textbf{92.89}  &   65.09   &  73.59  \\ \hline
\rowcolor[RGB]{222,235,247}
\textbf{\ourmodel}      &  90.13    &  92.21    &   \textbf{65.47}  &  \textbf{74.95}  \\  \hline
\end{tabular}
\end{center}
\end{table}

\section{Discussion}

Scene graph generation (SGG) in the operating room (OR) is essential for monitoring the surgical process, optimizing team collaboration, and improving workflow efficiency. 
Compared to reviewing surveillance records, generating scene graphs can highlight specific events of the surgical process and save the time cost of identifying abnormal events.
However, it is challenging to generate scene graphs directly from multimodal data streams.
Existing OR-SGG approaches~\cite{ozsoy20224d,ozsoy2023labrad} first detect human poses and objects, which are then used as inputs to predict scene graphs via GCNs.
% Although this multi-stage architecture reaches good performance, it is complicated and difficult to deploy in real clinical practices. 
% Driven by the above motivations, this paper aims to infer SGG in ORs using 2D multi-view images and 3D point cloud data without intermediate steps of processing.
\blue{Although this multi-stage architecture yields good performance, it requires stage-by-stage fine-tuning, which affects the overall effectiveness of end-to-end training of the model. Additionally, the multi-stage model is more cumbersome to deploy in real clinical practices, making it difficult to land in real-world surgery. 
Driven by the above motivations, this paper aims to infer scene graphs in ORs using 2D multi-view images and 3D point cloud data without intermediate steps of processing.}

We propose a novel single-stage multi-view bi-modal transformer (\ourmodel) model for SGG in ORs.
In practice, our model is constructed primarily based on a 2D SGG baseline augmented by point cloud samples.
In complex surgical procedures, leveraging point clouds to enhance geometric perception is crucial for prediction robustness.
Besides, the data streams from multiple cameras are also essential for monitoring specific actions of the head surgeon and overall scene recognition.
Therefore, our single-stage model includes a View-Sync Transfusion (VST) module for synthesizing multi-view semantic features and a Geometry-Visual Cohesion (GVC) operation for injecting structural cues extracted from point cloud features into synergic features.
To generate scene graphs in an end-to-end manner, we embed a relation-sensitive transformer in \ourmodel, which utilizes unified features and entity-aware embeddings to predict subject-object relations.
% We propose a novel single-stage multi-view bi-modal transformer (\ourmodel) model for SGG in ORs. 
% In practice, our model is constructed primarily based on a 2D SGG baseline augmented by point cloud samples.
% There are two main considerations for this design: (1) General 3D SGG benchmarks~\cite{wald2020learning,wang2023vl} mostly rely on given 3D object labels, where scene graphs are predicted by instance labeling combined with scene samples.
% In the real-world surgical procedure, however, it is difficult to have such prior knowledge of labels for humans, objects, and instruments, which need to predict scene graphs directly from overall scenes.
% (2) Most generic 2D SGG models~\cite{cong2023reltr,li2022sgtr,dong2022stacked,zheng2023prototype} employ an end-to-end framework that directly outputs triplets from images. 
% However, in complex surgical procedures, leveraging point clouds to enhance geometric perception is crucial for prediction robustness.
% Besides, the data streams from multiple cameras are also essential for monitoring specific actions of the head surgeon and overall scene recognition.
% Therefore, our single-stage model includes a View-Sync Transfusion (VST) module for synthesizing multi-view semantic features and a Geometry-Visual Cohesion (GVC) operation for injecting structural cues extracted from point features into synergic features.
% To generate scene graphs in an end-to-end manner, we embed a relation-sensitive transformer in \ourmodel, which utilizes unified features and entity-aware embeddings to predict subject-object relations.

As the only benchmark for OR SGG, the 4D-OR dataset~\cite{ozsoy20224d} contains a variety of interaction relations (\eg, Assisting, Drilling, and Touching) and spatial structural relations (\eg, Close-to and Lying-on). 
Unlike general 3D SGG datasets~\cite{wald2020learning} that focus primarily on spatial relations, 4D-OR requires more precise prediction of human-human or human-object interactions~\cite{kim2021hotr,liao2022gen,zhang2022exploring}. 
In this regard, our method is dedicated to the fine-grained extraction and fusion of multi-view semantic features, especially with our VST scheme and relational trait priors.
As shown in~\tabref{tab:sota1}, \ourmodel~achieves superior performance in predicting relationships initiated by surgeons, especially some subtle actions, such as Cementing, Cleaning, Cutting, and Sawing.
% Our proposed \ourmodel~not only can precept the operating room for knee surgery. Such capability of robustly predicting subtle relations benefits our method to be generalized to other types of surgeries.
Such capability to robustly predict subtle relationships benefits our model to be applied not only to knee surgery but also to be generalized to other types of surgery.
%It is important for real-world surgical workflow monitoring because models with robust relation predictions can be generalized to more different types of surgeries.
% Moreover, on the generic evaluation metrics (refer to~\tabref{tab:sota2}), our model still outperforms other state-of-the-art SGG models.
We believe our method can inspire more investigations on single-stage scene graph generation in operating rooms, further boosting the efficiency of surgical process analysis and promoting the development of surgical intelligence.

Although we contribute a single-stage multimodal OR-SGG framework, there are still several key points that deserve further exploration in this study.
% First, the accuracy of our model for entity detection can be further optimized. 
% As seen in~\figref{visual}, inaccurate entity classifications have an impact on the overall scene graphs.
First, we have not yet fully utilized the temporal information from multi-view video frames.
In future research, we will incorporate efficient temporal interaction operations into the single-stage pipeline to further improve temporal consistency.
Second, although the efficiency of our method has largely outperformed previous methods, \ie, 0.3s faster per scene, the proposed model has not yet reached real-time inference. 
In future efforts, we shall further improve the training and inference efficiency with the aid of pre-trained foundation models and a more lightweight design for cost-effective application in actual surgical scenes.
Furthermore, we consider integrating the clinical role labels directly into the SGG annotations, which will automatically recognize the attributes of each human node during scene graph generation, strengthening the fine-grained modeling for OR scenes.
We will optimize the above-mentioned issues and explore innovative techniques to further optimize the OR-SGG performance in future work.

\section{Conclusion}

We present a novel single-stage multi-view bi-modal transformer (\ourmodel) method for scene graph generation in operating rooms.
Specifically, we design a View-Sync Transfusion (VST) scheme to synthesize scene information from multiple viewpoints and a Geometry-Visual Cohesion (GVC) operation to implicitly merge 2D semantic features with 3D point cloud geometric cues.
Furthermore, a relation-sensitive transformer is proposed to directly predict scene graphs via dynamic relation queries and blend multimodal features.
The promising performance on the 4D-OR dataset demonstrates the effectiveness of our approach. 
More importantly, compared to existing multi-stage OR-SGG models, our single-stage end-to-end framework can generate scene graphs directly from surgical process content, greatly facilitating real-life applications in clinical surgical scenes.

% \section*{Acknowledgment}

% This work is supported by the following grants from

% \footnote{Place the actual footnote at the bottom of the column in which it is cited; do not put footnotes in the reference list (endnotes).}

% \begin{thebibliography}{00}
% \end{thebibliography}
\bibliographystyle{IEEEtran}
\bibliography{Reference}

\end{document}